\documentclass[10pt,twocolumn,letterpaper]{article}

\usepackage[pagenumbers]{iccv} %
\usepackage{amsmath}
\usepackage{booktabs}
\usepackage{caption}
\usepackage{subcaption}
\usepackage{enumitem}
\usepackage{multirow}
\usepackage{multicol}
\usepackage{cuted}
\usepackage{float}
\usepackage{pifont}
\usepackage{colortbl}

\newcommand{\cellfirst}{\cellcolor{Red!40}}
\newcommand{\cellsecond}{\cellcolor{Orange!25}}
\newcommand{\cellthird}{\cellcolor{Yellow!25}}

\newcommand{\cellred}{\cellcolor{Red!20}}

\newcommand{\cellgreendarkdark}{\cellcolor{Green!40}}
\newcommand{\cellgreendark}{\cellcolor{Green!30}}
\newcommand{\cellgreen}{\cellcolor{Green!20}}
\newcommand{\cellgreenlight}{\cellcolor{Green!10}}

\newcommand{\apref}[1]{Appx.~\ref*{#1}}
\newcommand{\secref}[1]{Sec.~\ref{#1}}
\newcommand{\tabref}[1]{Tab.~\ref{#1}}
\newcommand{\figref}[1]{Fig.~\ref{#1}}
\newcommand{\website}{\emph{Website}}
\newcommand{\supp}{\emph{Appendix}}

\newcommand{\method}{AV-Link\xspace}

\newcommand{\generator}{\mathcal{G}}
\newcommand{\generatoraudio}{\generator_{A}}
\newcommand{\generatorvideo}{\generator_{V}}
\newcommand{\generatoraudiotovideo}{\generator_{A2V}}
\newcommand{\generatorvideotoaudio}{\generator_{V2A}}
\newcommand{\multimodalblock}{\mathrm{FusionBlock}}

\newcommand{\token}{\boldsymbol{x}}
\newcommand{\activationvideo}{\boldsymbol{x}_{v}}
\newcommand{\activationaudio}{\boldsymbol{x}_{a}}
\newcommand{\projactivationvideo}{\bar{\boldsymbol{x}}_{v}}
\newcommand{\projactivationaudio}{\bar{\boldsymbol{x}}_{a}}
\newcommand{\activationvideomm}{\boldsymbol{\hat{x}}_{v}}
\newcommand{\activationaudiomm}{\boldsymbol{\hat{x}}_{a}}

\newcommand{\inputtensor}{\mathbf{X}}
\newcommand{\inputtensornoise}{\mathbf{X}_{0}}
\newcommand{\inputtensorclean}{\mathbf{X}_{1}}
\newcommand{\inputvideo}{\mathbf{V}}
\newcommand{\inputvideonoise}{\mathbf{V}_{0}}
\newcommand{\inputvideoclean}{\mathbf{V}_{1}}
\newcommand{\inputaudio}{\mathbf{A}}
\newcommand{\inputaudionoise}{\mathbf{A}_{0}}
\newcommand{\inputaudioclean}{\mathbf{A}_{1}}

\newcommand{\velocity}{{v}}

\newcommand{\datasetaudio}{\mathcal{D}_{A}}

\newcommand{\datasetaudiovideo}{\mathcal{D}_{AV}}

\newcommand{\temporalalignmentfunction}{\tau}
\newcommand{\datadistribution}{p_{d}}
\newcommand{\timestepdistribution}{p_{t}}
\newcommand{\timestepdistributionaudio}{p_{t_a}}
\newcommand{\timestepdistributionvideo}{p_{t_v}}
\newcommand{\noisedistribution}{p_{n}}
\newcommand{\timeindexrope}{n}
\newcommand{\difftimestep}{t}

\newcommand{\difftimestepaudio}{t_{a}}
\newcommand{\difftimestepvideo}{t_{v}}

\newcommand{\numaudiotokens}{T_{a}}
\newcommand{\numvideotokens}{T_{v}}
\newcommand{\numaudioaetokens}{T_{a}}
\newcommand{\numvideoaetokens}{T_{v}}
\newcommand{\numchannels}{D}
\newcommand{\numaudiochannels}{D_{a}}
\newcommand{\numvideochannels}{D_{v}}
\newcommand{\numaudioaechannels}{D_{a}}

\newcommand\nnfootnote[1]{%
  \begin{NoHyper}
  \renewcommand\thefootnote{*} %
  \footnotetext{#1}%
  \renewcommand\thefootnote{\arabic{footnote}} %
  \addtocounter{footnote}{-1}%
  \end{NoHyper}
}

\definecolor{iccvblue}{rgb}{0.21,0.49,0.74}
\usepackage[pagebackref,breaklinks,colorlinks,allcolors=iccvblue]{hyperref}

\def\paperID{9100} %
\def\confName{ICCV}
\def\confYear{2025}

\title{\method: Temporally-Aligned Diffusion Features for Cross-Modal \\ Audio-Video Generation}
\author{%
  \makebox[\textwidth][c]{%
    \text{Moayed Haji-Ali}$^{1,2,*}$ \qquad
    \text{Willi Menapace}$^{2}$ \qquad
    \text{Aliaksandr Siarohin}$^{2}$ \qquad
    \text{Ivan Skorokhodov}$^{2}$ 
  } \and
  \vspace{2em}
  \makebox[\textwidth][c]{%
    \text{Alper Canberk}$^{2}$ \qquad
    \text{Kwot Sin Lee}$^{2}$ \qquad
    \text{Vicente Ordonez}$^{1}$ \qquad
    \text{Sergey Tulyakov}$^{2}$ 
  } \\
  \vspace{2em}
  \centering
  \parbox{\linewidth}{\centering
    ${}^1\text{Rice University}$ \qquad \qquad ${}^2\text{Snap Inc}$\\
    \vspace{1em}
    Project Webpage: \href{https://snap-research.github.io/AVLink}{\color{blue}https://snap-research.github.io/AVLink}
  }
}
\begin{document}
\twocolumn[{%
\renewcommand\twocolumn[1][]{#1}%
\maketitle
\begin{center}
    \centering
    \captionsetup{type=figure}
    \vspace{-0.4in}
    \includegraphics[width=\textwidth]{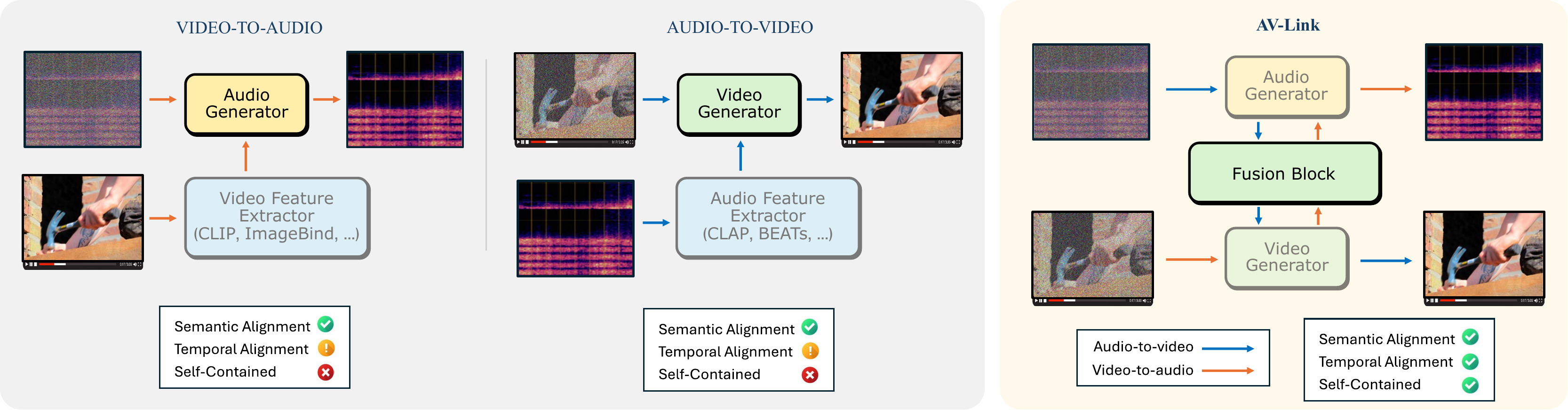}
    \captionof{figure}{Compared to current Video-to-Audio and Audio-to-Video methods, \method{} provides a unified framework for these two tasks. Rather than relying on feature extractors pretrained for other tasks (e.g.~CLIP~\cite{clip}, CLAP~\cite{clap}), we directly leverage the activations from pretrained frozen Flow Matching models using a Fusion Block to achieve precise time alignment between modalities. Our approach offers competitive semantic alignment and improved temporal alignment in a self-contained framework for both modalities.}
      \label{fig:teaser}
\end{center}%
}]

\begin{abstract}
\nnfootnote{Work mainly done during an internship at Snap Inc.}
We propose \method, a unified framework for Video-to-Audio (A2V) and Audio-to-Video (A2V) generation that leverages the activations of \emph{frozen} video and audio diffusion models for temporally-aligned cross-modal conditioning. The key to our framework is a Fusion Block that facilitates bidirectional information exchange between video and audio diffusion models through temporally-aligned self attention operations. Unlike prior work that uses dedicated models for A2V and V2A tasks and relies on pretrained feature extractors, \mbox{\method} achieves both tasks in a single self-contained framework, directly leveraging features obtained by the complementary modality (i.e. video features to generate audio, or audio features to generate video). Extensive automatic and subjective evaluations demonstrate that our method achieves a substantial improvement in audio-video synchronization, outperforming more expensive baselines such as the MovieGen video-to-audio model.

\end{abstract}
\vspace{-3mm}
    
\section{Introduction}

Generative models are becoming proficient at synthesizing high-quality video~\cite{sora,villegas2022phenaki,snapvideo,make_a_video,videofusion,pyramidalflow,vidstyleode} and audio~\cite{genau,stableaudio,audioldm,audioldm2,audiobox}.
Given the intrinsic correlation between audio and video, there has been a growing interest in cross-modal generation including Video-to-Audio (V2A) and Audio-to-Video (A2V). The goal of V2A models is to add sound that is both semantically and temporally aligned to video content~\cite{difffoley,foleycrafter,seeingandhearing,moviegen}. Conversely, A2V models~\cite{tempotoken,audiosynchronizedvisualanimation, lee2023aadiffaudioalignedvideosynthesis} seek to generate video that plausibly represent the sounds of a given audio. Since text has been established as a robust conditioning modality to guide the semantics of generated audio and video~\cite{cogvideox, huang2023makeanaudio2}, achieving accurate audio–video temporal alignment remains as the \emph{most critical} and challenging aspect in cross-modal generation. Current V2A models~\cite{difffoley,foleycrafter,seeingandhearing,moviegen} typically depend on pre-trained visual feature extractors such as~CLIP~\cite{clip}, MetaCLIP~\cite{metaclip}, ImageBind~\cite{imagebind}, and CAVP~\cite{difffoley} to condition audio generation and create soundscapes that are semantically aligned with visual frames. Similarly, A2V methods condition video generators on features from pretrained audio models such as ImageBind~\cite{imagebind}, and BEATs~\cite{beats}. However, the reliance on modality-specific feature encoders necessitates distinct models for V2A and A2V, limiting the potential for a unified framework and 
complicating the deployment for applications requiring comprehensive capabilities: text-to-audio (T2A), text-to-video (T2V), V2A, and A2V.

Recognizing the advantages of a single framework capable of generating both audio and video, recent work explores training a single model for joint audio-video generation~\cite{mmdiffusion,avdit,versatilediffusiontransformermixture,simplestrongbaselinesounding}. Although promising, these models have a performance gap with respect to conditional models (\ie dedicated V2A and A2V models), and often operate on simpler datasets such as landscape videos~\cite{mmdiffusion, avdit, lee2022soundguidedsemanticvideogeneration} or exhibit weaker audio-video synchronization~\cite{simplestrongbaselinesounding}.

In this work, we introduce \method{}, a unified framework that is capable of \emph{semantically} and \emph{temporally} aligned generation for both V2A and A2V as shown in \figref{fig:teaser}. Starting from two well-trained and \emph{frozen} flow models for audio and video generation, we propose to use their activations as conditioning signals for V2A or A2V synthesis, bypassing the need for pretrained feature extractors. Such activations contain rich semantics~\cite{hyperfeatures}, but they are also \emph{temporally-aligned} since these activations are trained to generate modalities with a temporal component (\ie audio and video). This is motivated by prior findings showing that activations in image diffusion models encode substantial semantic and spatial information, making them useful for pixel-aligned tasks such as semantic keypoint matching~\cite{hyperfeatures,hedlin2023unsupervisedsemanticcorrespondenceusing} segmentation~\cite{yang2023diffusionmodelrepresentationlearner} and depth estimation~\cite{ke2023repurposing}.

To preserve the quality of the 
frozen
unimodal 
generators while enabling cross-modal generation, we introduce a Fusion Block that connects the audio and video modalities by conditioning the generation of one modality on the activations from the other. Furthermore, we propose time-aligned Rotary Position Embedding (RoPE) as a flexible mechanism to align audio and video tokens across the temporal axis, demonstrating its benefits over prior approaches for temporal alignment.
Moreover, although static diffusion model activations serve as a robust conditioning signal, cross-modal generation may benefit from a conditioning signal that dynamically evolves throughout the generation process. To this end, we introduce a \emph{symmetric feature reinjection} strategy that leverages the diffusion prior of the pretrained model for the conditioning modality to refine the conditioning signal, enabling bidirectional information exchange between the conditioning and generated modality. 
By introducing a relatively small set of parameters (186M parameters), \method{} offers a compact design that simplifies training and deployment for applications requiring comprehensive T2V, T2A, V2A, and A2V generation. Rather than focusing on improving single-modality generation—a topic extensively explored in concurrent studies~\cite{genau, cogvideox}—this work builds on those efforts by unifying pretrained audio and video generators in a single framework, emphasizing temporal alignment as the core challenge in cross audio-visual generation.

In summary, we propose \method{}, a unified framework to address A2V and V2A tasks. We show for the first time that video and audio diffusion activations contain rich temporal information and can replace specialized pretrained feature extractors while improving \emph{temporal} alignment. The key to our method is a novel design of a Fusion Block that enables bidirectional information exchange between the conditioning and generated modalities through a symmetric feature reinjection mechanism. When evaluated against the state-of-the-art V2A method on VGGSounds~\cite{chen2020vggsound}, our method shows the best overall performance and drastically improves temporal alignment as measured by Onset ACC~\cite{conditionalgenerationaudiovideo} by up to $76.4\%$ over the best baseline. User studies show a strong preference for our method which is preferred over Movie Gen Audio~\cite{moviegen} (13B parameters) $63.6\%$ of the time with regards to temporal alignment. In the A2V task, \method{} surpasses TempoTokens~\cite{temporallyalignedaudiovideo} in video quality and audio-video alignment, with user studies showing an overwhelming preference for our method.

\section{Related Work}
\label{sec:related}

\noindent\textbf{Text-to-Audio Generation}
Text-to-audio methods generate audio conditioned a given text~\cite{tango2,auffusion,stableaudio,audioldm2,huang2023makeanaudio2,genau}. Recent works employ scalable transformer-based diffusion backbones over a latent representation of Mel-spectrograms to achieve superior quality. Make-an-Audio 2~\cite{huang2023makeanaudio2} encodes the spectrogram with a 1D VAE and uses a DiT~\cite{dit} backbone. GenAU~\cite{genau} introduced a scalable dataset collection framework and transformer architecture to produce high-quality sound effects. Movie Gen~\cite{moviegen} directly encodes waveforms with a VAE, bypassing intermediate Mel-spectrogram representations. Such approaches operate in a text-conditioned setting \cite{stableaudio, audioldm2, huang2023makeanaudio2}, and showed the importance of text encoding in achieving better quality and text-audio alignment. 
Our work focuses on establishing a unified framework for text-conditioned V2A and A2V, posing a more complex challenge due to the need for precise temporal alignment with the conditioning modality.

\noindent\textbf{Video-to-Audio Generation} 
The task of generating audio from silent videos recently emerged as a difficult generative task~\cite{TiVA,lovalongformvideotoaudiogeneration,maskedgenerativevideotoaudiotransformers, mei2023foleygen, mmldmmultimodallatentdiffusion, videofoleytwostagevideotosoundgeneration, mao2024tavgbench, egosonicsgeneratingsynchronizedaudio, sonicdiffusionaudiodrivenimagegeneration, drawaudioleveragingmultiinstruction, semanticallyconsistentvideotoaudiogeneration, action2soundambientawaregenerationaction, gottahearallsound, videoguidedfoleysoundgeneration, vintagejointvideotext, lin2024vmasvideotomusicgenerationsemantic, muvivideotomusicgenerationsemantic, momudiffusionlearninglongtermmotionmusic, motiontodance} and gained relevance not only for its potential in simplifying movie production but also because of the recent development of high-quality video generation methods which produce no audio output~\cite{snapvideo,moviegen,cogvideox,pyramidalflow,t2vturbov2,allegro, sora, svd, videoldm}.
Video feature representations play a crucial role in achieving highly aligned generation and have been the focus of most recent work~\cite{frieren,specvqgan,im2wav,sonicvision,wang2023v2amapper,difffoley,foleycrafter,moviegen,ren2024stav2avideotoaudiogenerationsemantic,huang2024rhythmicfoleyframeworkseamless}. SpecVQGAN~\cite{specvqgan} employs a cross-modal transformer conditioned on ImageNet-pretrained visual backbones. 
V2A-Mapper~\cite{wang2023v2amapper} aligns CLIP~\cite{clip} and CLAP~\cite{clap} to leverage a pretrained text-to-audio model for V2A. Diff-Foley~\cite{difffoley} and Frieren~\cite{frieren} address alignment using CAVP~\cite{cavp}, an ad-hoc contrastive video-audio representation. FoleyCrafter~\cite{foleycrafter} augments a frozen audio generator with a learnable semantic adapter and temporal controller. Movie~Gen~Audio~\cite{moviegen} obtains strong video-audio alignment by conditioning on a descriptive text caption and per-frame MetaCLIP~\cite{metaclip} features. Concurrent work~\cite{tamingmultimodal,multifoleypaper} demonstrated remarkable capabilities by leveraging training on audio-only datasets. Despite such advancements, the unification of A2V and V2A in a single framework remains a challenge.
We show that temporally-aligned V2A and A2V generation can be achieved within the same framework by leveraging existing diffusion activations without the need for specialized feature extractors.

\noindent\textbf{Audio-to-Video Generation}
Similar to the Video-to-Audio task, recent work has focused on building suitable audio representations, paying special attention to time alignment.
\emph{Lee}~\etal~\cite{lee2022soundguidedsemanticvideogeneration} employ a Sound Inversion Encoder to map audio features to the StyleGAN latent space. Seeing-and-Hearing~\cite{seeingandhearing} propose a video-audio aligner based on ImageBind~\cite{imagebind}. 
TPoS~\cite{jeong2023powersoundtposaudio} manipulates generated images conditioned on audio representations extracted through an ad-hoc temporally-aligned audio encoder.  
TempoTokens~\cite{tempotoken} aligns a frozen video generator to the audio modality by employing features from a finetuned BEATs encoder. AADif~\cite{lee2023aadiffaudioalignedvideosynthesis} leverages the CLAP~\cite{clap} audio encoder and audio magnitude to condition generation. Recently, AVSyncD~\cite{audiosynchronizedvisualanimation} encodes audio into temporally-dependent tokens using ImageBind~\cite{imagebind}. We show that using activations from a pretrained audio generator provides a stronger conditioning signal than external feature extractors, yielding improved temporal alignment. 

\noindent\textbf{Joint Video-Audio Generation}
Several works emerged that generate the audio and video modalities jointly.~\cite{discriminatorguidedcooperativediffusionjoint,versatilediffusiontransformermixture, cmmdcontrastivemultimodaldiffusion, languageguidedjointaudiovisualediting, sequentialcontrastiveaudiovisuallearning,visionaudiobeyondunified} MM-Diffusion~\cite{mmdiffusion} proposes a diffusion framework based on a Coupled U-Net for simultaneously denoising the audio and video streams.
Seeing-and-Hearing~\cite{seeingandhearing} employs ImageBind~\cite{imagebind} to establish alignment between modalities. AVDiT~\cite{avdit} introduces a multimodal DiT design and learns multiple conditional distributions over audio and video through a mixture of noise levels, enabling V2A, A2V, and joint audio-video generation. Such approaches, however, present lower performance than conditional methods~\cite{foleycrafter,moviegen}. Additionally, they are trained from scratch on audio-video datasets. Since most large-scale audio and video datasets are curated around a single modality, the other modality may be missing, have insufficient quality, or unlabeled~\cite{chen2024panda70m, genau}, this limits the data scalability of such methods. Thus, in this work, we target cross-modal generation (\ie V2A and A2V) rather than joint audio-video generation.

\vspace{-0.5em}
\section{Method}
\vspace{-0.2em}
\begin{figure}[t]
\centering
\includegraphics[width=0.63\columnwidth]{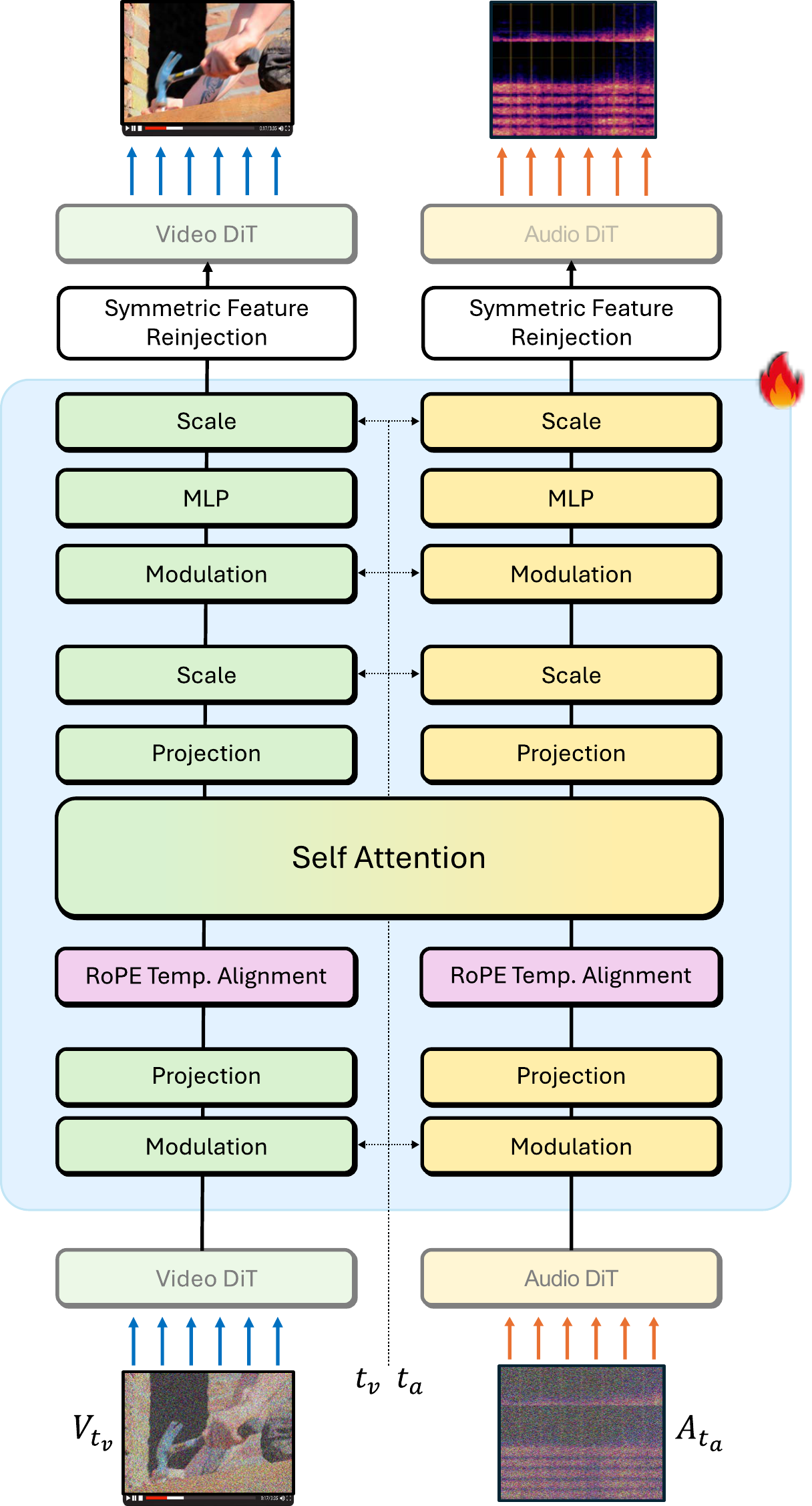}
  \caption{Design of the proposed Fusion Block connecting the frozen video and audio backbones. A RoPE-based temporal alignment mechanism aligns the representation of the two modalities which are processed by self attention. Video and audio features are symmetrically reinjected into the frozen generators. The block is regularly applied multiple times throughout the backbones.} 
  \vspace{-2mm}
  \label{fig:architecture}
\end{figure}

Starting with pretrained and \emph{frozen} Text-to-Audio and Text-to-Video generators, we aim to combine the two under a unified architecture to produce Video-to-Audio $\generatorvideotoaudio$ and Audio-to-Video $\generatoraudiotovideo$ generators. To achieve this, we propose \method{}, a symmetric architecture that benefits from the activations of one generator to condition the generator of the other modality.
In \secref{sec:preliminaries}, we describe the background on Flow Matching models. We then describe in \secref{sec:audio_model}, the architecture of our pretrained audio and video generators. Finally, in \secref{sec:modality_fusion}, we describe the architecture of the multimodal Fusion Blocks that connect the two generators. 

\subsection{Background}
\label{sec:preliminaries}

We base our generative models on the Flow Matching framework \cite{lipman2023flowmatching,liu2022rectifiedflow}. Flow Matching expresses generation of data $\inputtensorclean \sim \datadistribution$ as the progressive transformation of $\inputtensornoise$ following a path connecting samples from the two distributions. In its simplest formulation~\cite{liu2022rectifiedflow}, the path is instantiated as a linear interpolation between the samples:
\begin{equation}
\inputtensor_\difftimestep = \difftimestep \inputtensorclean + (1-\difftimestep) \inputtensornoise
\text{,}
\end{equation}
and $\inputtensornoise \sim \noisedistribution=\mathcal{N}(0,\mathbb{I})$ originate from a noise distribution. 

We can move along the path following the velocity $\velocity_\difftimestep = \frac{d\inputtensor_\difftimestep}{d\difftimestep} = \inputtensorclean - \inputtensornoise$ approximated by learnable $\generator$ minimizing:
\begin{equation}
\label{eq:rf}
    \mathcal{L} = \mathbb{E}_{\difftimestep\sim\timestepdistribution, \inputtensorclean\sim\datadistribution, \inputtensornoise\sim\noisedistribution} ~\big\lVert \generator(\inputtensor_{\difftimestep}, \difftimestep) - \velocity_\difftimestep \big\rVert^2_2
    \text{,}
\end{equation}
where $\timestepdistribution$ indicates a training distribution over time $\difftimestep$, which we instantiate as a logit normal distribution~\cite{sd3}. At inference time, an ODE solver such as a first-order Euler solver can be employed to produce samples $\inputtensorclean$ starting from Gaussian noise $\inputtensornoise$ using the model velocity estimates.

\subsection{Base Models}
\label{sec:audio_model}
\textbf{Audio Model.} Given an audio signal from audio dataset $\datasetaudio$, we produce a corresponding Mel-spectrogram and adopt a 1D-VAE~\cite{genau} to encode it into a 1D sequence of latent representations. This latent audio representation $\inputaudioclean$ thus has shape $\mathbb{R}^{\numaudioaetokens \times \numaudioaechannels}$, where $\numaudioaetokens$ and $\numaudioaechannels$ respectively indicate the number of tokens and their dimensionality produced by the autoencoder.
The sequence is 
modeled %
by a DiT~\cite{dit} (see \figref{fig:architecture}) consisting of a stack of identical blocks containing a self attention operation followed by an MLP. Information on the current timestep $\difftimestep$ is injected through an adaptive 
LayerNorm~\cite{perez2018film} %
and the attention operator is augmented with 1D-RoPE~\cite{rope} to encode positional information. 
To condition on text prompts, we encode them using the T5~\cite{raffel2022exploring} encoder and insert a cross attention layer attending to the text embeddings after each self attention operation~\cite{moviegen}. The model contains 576M parameters and is trained to generate variable-length audio clips at 16kHz following previous work~\cite{huang2023makeanaudio2, audioldm2}. We train it using the Flow Matching objective (Eq.~\eqref{eq:rf}) over 100k iterations, with a learning rate of $3e\!-\!4$ and a batch size of 1024, distributed across 8 A100 GPUs. During inference, a pre-trained Vocoder \cite{kong2020hifigan} converts the generated Mel-spectrograms into waveforms.

\noindent\textbf{Video Model.} Given an RGB video $\inputvideoclean$, we flatten it to shape $\mathbb{R}^{\numvideoaetokens \times 3}$, where $\numvideoaetokens$ indicates the total number of video pixels.
We adopt a DiT architecture symmetric to the audio model, and use 3D-RoPE~\cite{chu2024visionllama}. 
Due to the computational costs associated with large scale video generators~\cite{cogvideox, moviegen} and the orthogonality between improvement of single-modality generation and \method{}, we adopt a relatively small model (576M parameters) for our experiments, enabling efficient training and inference on the video modality.
We found that when using a latent diffusion formulation, its motion and structural quality do not match the ones from an RGB model of the same size. Consequently, we opt to train a base video model to generate 5.16-second videos in the RGB space at a resolution of 36 x 64 pixels and 6 frames-per-second (FPS). The resolution is later increased to 512 x 288 at 24 FPS through a latent video upsampler operating in the latent space of a MAGVIT-v2~\cite{yu2024languagemodelbeatsdiffusion} autoencoder that is directly finetuned from the video model. 
The base video model is trained for 250k iterations with a batch size of 512, while the upsampler uses a batch size of 384 for 25k iterations. Both models are trained with a learning rate of $3e\!-\!4$ on 16 A100 GPUs.

\subsection{Multimodal Fusion Block}
\label{sec:modality_fusion}

Cross-modal generation requires a high-quality representation for the conditioning modality~\cite{difffoley,foleycrafter}. In the context of V2A, obtaining high-quality video features has been the focus of recent work, which explored the usage of contrastive representations such as ImageBind~\cite{seeingandhearing}, CLIP~\cite{foleycrafter}, MetaCLIP~\cite{moviegen}, or built ad hoc video representations~\cite{difffoley, foleycrafter}. While such representations capture the overall video semantic, we observe that they lack precise audio-video temporal alignment, producing characteristic temporal misalignment (see \secref{sec:ablations}). 
On the other hand, the video generator is capable of generating videos from scratch, implying that its activations contain semantically and temporally aligned information that can be leveraged for V2A.
We thus propose a fully symmetric framework for V2A and A2V that leverages activations from the frozen generator of the conditioning modality to achieve aligned cross-modal generation using no ad-hoc feature extractors.

\noindent\textbf{Fusion Blocks} We address V2A and A2V by linking frozen audio and video generators $\generatoraudio$, $\generatorvideo$ through the Fusion Block depicted in \figref{fig:architecture}.
Consider the output activations of audio and video DiT blocks $\activationaudio \in \mathbb{R}^{\numaudiotokens \times \numaudiochannels}$, $\activationvideo \in \mathbb{R}^{\numvideotokens \times \numvideochannels}$ at some diffusion timestep $\difftimestepaudio$, $\difftimestepvideo$. We derive the activation of the conditional modality from a ground truth input (\eg video for the V2A task). Initially, we project activations into a common dimension $\numchannels$ to produce $\projactivationaudio \in \mathbb{R}^{\numaudiotokens \times \numchannels}$, $\projactivationvideo \in \mathbb{R}^{\numvideotokens \times \numchannels}$ respectively. A multi-head self attention operation is then applied to the concatenation of audio and video features $\{ \projactivationaudio, \projactivationvideo\}$. Finally, an MLP projects the output back to the original dimensionality, producing  video-aware audio activations $\activationaudiomm \in \mathbb{R}^{\numaudiotokens \times \numaudiochannels}$ and audio-aware video activations $\activationvideomm \in \mathbb{R}^{\numvideotokens \times \numvideochannels}$ as: 
\begin{equation}
    \activationaudiomm, \activationvideomm = \multimodalblock(\activationaudio,\activationvideo,\difftimestepaudio,\difftimestepvideo)
    \text{.}
\end{equation}
The Fusion Block presents a simple and symmetric design that equally treats the different modalities to condition the other modality.

\noindent\textbf{Symmetric Feature Reinjection} 
While the Fusion blocks effectively transfer the conditioning signal to the generated modality, we hypothesize that this signal can benefit from a continuous refinement driven by the generated modality. The symmetric architecture of the Fusion Blocks enables bidirectional information flow between the conditioning and generated modalities, rather than a one-way flow from the conditioning to the generated modality. 
Therefore, we propose to pass the conditional-aware (\ie $\activationaudiomm$ in the V2A task) and generation-aware activations (\ie $\activationvideomm$ in V2A task) to subsequent DiT blocks within the respective frozen backbones, thereby injecting cross-modal information.
In contrast to using static feature extractors~\cite{moviegen, difffoley, foleycrafter}, our design, which we call \emph{symmetric feature reinjection}, dynamically improves the quality of the activations extracted from successive blocks of the conditioning modality.

\noindent\textbf{Multimodal Temporal Alignment} To facilitate temporal alignment between modalities, we use temporally-aligned 1D RoPE~\cite{rope} embeddings in the self attention operation. We express 1D temporal RoPE as:
\begin{equation}
    \mathrm{RoPE}(\token_n,\temporalalignmentfunction(\timeindexrope))=\token_\timeindexrope e^{i\temporalalignmentfunction(\timeindexrope)\theta_\mathrm{base}}
    \text{,}
\end{equation}
where $\token$ represents an audio or video token for a simplified two-channel case, $\timeindexrope$ represents its index on the temporal axis, $\theta_\mathrm{base}$ is the base frequency, and $\temporalalignmentfunction$ is a temporal alignment function defined as: 
\[
\temporalalignmentfunction(\timeindexrope)=
\begin{cases}
     \timeindexrope, & \text{if } \token \text{~is a video token}\\
     \timeindexrope \frac{\eta_v}{\eta_a} , & \text{if } \token \text{~is an audio token}
\end{cases}
\]
Where $\eta_a$, $\eta_v$ are the number of tokens to represent one second of audio and video respectively. 
Intuitively, this approach rotates corresponding audio and video tokens by the same angle proportional to their temporal position, thus establishing temporal correspondence that facilitates temporally-synced multimodal information exchange. Crucially, this design aligns the modalities within the inserted Fusion Blocks while keeping the pretrained generators frozen. We show in~\secref{sec:ablations} that this design is favored over previous temporal alignment approaches.

\noindent\textbf{Time-Aware Feature Fusion} We inject the information on the flow timesteps for audio $\difftimestepaudio$ and video $\difftimestepvideo$ through adaptive adaLN-Zero~\cite{dit}, with audio and video parameter sets receiving information on both timesteps. Such design creates conditioning features that are most suited for the current flow time for the generated modality, increasing the effectiveness of \emph{symmetric feature reinjection}.

\noindent\textbf{Training} We train the Fusion Block on a paired audio-video dataset $\datasetaudiovideo$ keeping both generators \emph{frozen} and using the rectified flow objective. For the case of V2A, we use the training objective:
\begin{multline}
    \resizebox{0.85\columnwidth}{!}{$%
    \mathcal{L} = \mathbb{E}_{\difftimestepaudio, \difftimestepvideo, \inputaudioclean,\inputvideoclean, \inputaudionoise} ~\big\lVert \generatorvideotoaudio(\inputaudio_{\difftimestepaudio}, \inputvideo_{\difftimestepvideo}, \difftimestepaudio, \difftimestepvideo) - (\inputaudioclean - \inputaudionoise) \big\rVert^2_2$}
    \text{,}\\
    \resizebox{0.65\columnwidth}{!}{$%
    \difftimestepaudio\sim\timestepdistributionaudio, \difftimestepvideo\sim\timestepdistributionvideo, \inputaudioclean,\inputvideoclean\sim\datasetaudiovideo, \inputaudionoise\sim\noisedistribution$}
    \text{,}
\end{multline}
and use a fully symmetric one for A2V. Keeping the audio and video generators frozen, we train the Fusion Blocks for 50k iterations with a learning rate of $3e\!-\!4$ and a batch size of 256 for the V2A and A2V tasks. Additionally, we consider using separate parameter sets for the V2A and A2V tasks or training them jointly (see \secref{sec:ablations}). 
We discuss architectural details in \apref{sec:architecture_details}, and training details in \apref{sec:training_details}.

\noindent\textbf{Inference} In the V2A setting, we consider the input conditioning video $\inputvideoclean$ and compute partially noised video $\inputvideo_{\difftimestepvideo}$ at the chosen flow time $\difftimestepvideo$, which we set to a fixed level during the entire inference phase to serve as input for the video backbone. As we later show in \ref{sec:ablations}, it is crucial to inject some noise into the conditioning modality to extract useful features from the conditioning model. We sample $\inputaudionoise$ from Gaussian noise at $\difftimestepaudio\!=\!0$ and use velocity predictions from $\generatorvideotoaudio$ to progressively denoise it into $\inputaudioclean$. We perform sampling using a first-order Euler solver with 64 sampling steps. Text prompts can optionally be used for both modalities to condition generation. We perform A2V inference completely symmetrically. 

\begin{figure}
    \centering
    \includegraphics[width=0.99\linewidth,clip]{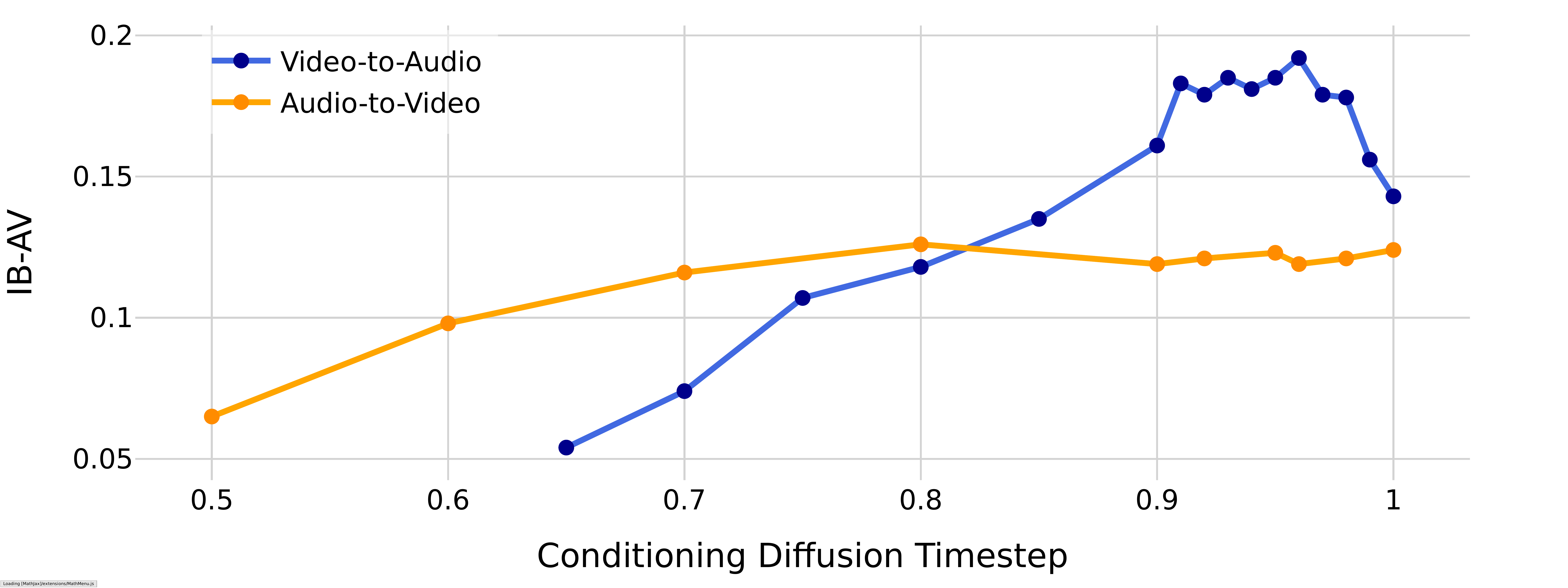}
    \caption{Visualization of Audio-to-Video and Video-to-Audio generation performance for various value of flow timesteps $\difftimestep$ for conditioning features. Best performance is achieved when conditioning features are close to be fully denoised, \ie $\difftimestep \in [0.8, 0.98]$. }
    \label{fig:video_time_choice}
\end{figure}

\section{Experiments}

\begin{table}[t!]
\begin{center}
\setlength{\tabcolsep}{3.0pt}
\resizebox{\linewidth}{!}{%
\begin{tabular}{l c c c c c c c c }

\toprule

 & Prompt & FAD $\downarrow$ & FD $\downarrow$ & CLAP $\uparrow$ & IS $\uparrow$ & IB-AI $\uparrow$ & IB-AV $\uparrow$ & Ons. ACC $\uparrow$ \\

\midrule

$\text{Diff-Foley}^*$~\cite{difffoley}                  &  & 11.00 & 28.71 & - & 7.88 & 0.115 & 0.121 & 0.140 \\
$\text{S\&H}^*$~\cite{seeingandhearing}   &  & 29.22 & 66.51 & - & 2.09 & 0.179 & 0.189 & 0.128  \\
$\text{FoleyCrafter}^*$~\cite{foleycrafter}             &    & 4.62 & 18.66 & - & 9.10 &  \cellthird{0.204} & \cellsecond{0.215} & 0.285 \\
$\text{Frieren}^\dag$~\cite{frieren}            &    & 3.52 & 16.50 & - & 7.78 & 0.178 & 0.184 & 0.301\\
$\text{V2A-Mapper}^\dag$~\cite{wang2023v2amapper}            &   & 2.77 & 16.60 & - & 7.65 &  0.177 & 0.183 & 0.281 \\
Ours (VGGSounds)                             &   &  \cellsecond{2.02} & \cellsecond{13.68}  & - & \cellfirst{10.06} & 0.203 & \cellthird{0.214} & \cellthird{0.405}   \\
Ours-Joint                                  &   &  \cellthird{2.19} & \cellfirst{13.15}  & - & \cellthird{9.24} & \cellsecond{0.205} & \cellsecond{0.215} & \cellsecond{0.409}   \\ 

Ours                                         &   &  \cellfirst{1.58} & \cellthird{14.17} & - & \cellsecond{9.93} & \cellfirst{0.207} & \cellfirst{0.223} & \cellfirst{0.531}  \\
\midrule
\midrule

$\text{S\&H}^*$~\cite{seeingandhearing}  & \checkmark   & 10.72 &  25.44 & 0.186 & 6.06 & \cellfirst{0.30} & \cellfirst{0.318} & 0.08  \\
$\text{FoleyCrafter}^*$~\cite{foleycrafter}            & \checkmark   & 2.99 & 18.05 & 0.212 & 10.88 & 0.208 & 0.219 & 0.307  \\

Ours (VGGSounds)                             &  \checkmark &  \cellsecond{1.91} & \cellsecond{13.19}  & \cellfirst{0.236} & \cellfirst{13.57} & \cellsecond{0.225} & \cellsecond{0.238} & \cellthird{0.429}   \\

Ours-Joint                                 & \checkmark  & \cellthird{2.23} & \cellthird{13.74} & \cellthird{0.224} & \cellsecond{13.44} & 0.212 & 0.224 & \cellsecond{0.471}  \\
Ours                                        & \checkmark   & \cellfirst{1.33} & \cellfirst{11.99} & \cellsecond{0.228} & \cellthird{12.40} & \cellthird{0.214} & \cellthird{0.226} & \cellfirst{0.540} \\

\bottomrule

\end{tabular}
}
\end{center}
\caption{\label{tab:baselines_comparison_v2a} Evaluation results on the V2A task on the VGGSounds~\cite{chen2020vggsound} benchmark. * evaluated using the released code and checkpoints, $\dag$ evaluated using results obtained directly from the authors. We report a variant of our method when training on VGGSounds only and when jointly train on A2V and V2A.}
\end{table}

\begin{table*}[ht]
\begin{center}
\setlength{\tabcolsep}{3.0pt}
\footnotesize
\resizebox{0.75\linewidth}{!}{%
\begin{tabular}{l c c c c | c c c c c c c }

\toprule

 & CLAP $\uparrow$ & IS $\uparrow$ & IB-AI $\uparrow$ & IB-AV $\uparrow$ & FAD $\downarrow$ & FD $\downarrow$ & CLAP $\uparrow$ & IS $\uparrow$ & IB-AI $\uparrow$ & IB-AV $\uparrow$ & Ons. ACC $\uparrow$ \\

\midrule
& \multicolumn{4}{c}{\emph{Movie Gen Benchmark}} & \multicolumn{7}{c}{\emph{VGGSounds}}\\
\midrule
\emph{Conditioning timestep $\difftimestep$:} &&&&&& \\
- Uniform samp. & \cellfirst{0.216} & 4.19 & 0.103 & 0.111 & 6.91 & 27.33 & 0.108 & 8.12 & 0.180 & 0.190 & 0.413  \\
- Fixed (ours)  & 0.192 & \cellfirst{6.53} & \cellfirst{0.150} & \cellfirst{0.155}  & \cellfirst{4.79} & \cellfirst{18.91} & \cellfirst{0.131} & \cellfirst{9.21} & \cellfirst{0.210} & \cellfirst{0.222} & \cellfirst{0.415} \\
\midrule

\emph{Conditioning features type:} &&&& \\
- CAVP 
  & \cellsecond{0.184} 
  & \cellthird{4.62} 
  & {0.116} 
  & {0.120} 
  & \cellthird{3.63} 
  & 24.36 
  & \cellthird{0.098} 
  & 7.37 
  & 0.172 
  & 0.180 
  & \cellthird{0.383} 
\\
- CAVP w/FT 
  & 0.169 
  & \cellsecond{5.69} 
  & \cellthird{0.136} 
  & \cellthird{0.143} 
  & \cellsecond{3.33} 
  & \cellthird{23.81} 
  & \cellthird{0.098} 
  & \cellsecond{9.18} 
  & \cellthird{0.197} 
  & \cellthird{0.208} 
  & 0.371 
\\
- CLIP 
  & \cellthird{0.171} 
  & 3.26 
  & \cellsecond{0.143} 
  & \cellsecond{0.150} 
  & \cellfirst{2.49} 
  & \cellsecond{21.47} 
  & \cellsecond{0.117} 
  & \cellthird{8.56} 
  & \cellfirst{0.234} 
  & \cellfirst{0.247} 
  & \cellsecond{0.386} 
\\
- Diffusion features (ours) 
  & \cellfirst{0.192} 
  & \cellfirst{6.53} 
  & \cellfirst{0.150}
  & \cellfirst{0.155}
  & 4.79 
  & \cellfirst{18.91} 
  & \cellfirst{ 0.131}
  & \cellfirst{9.21} 
  & \cellsecond{0.210} 
  & \cellsecond{0.222} 
  & \cellfirst{0.415} 
\\

\midrule
\emph{Fusion block arrangement:} &&&& \\
- After Block-1      & \cellthird{0.170} & \cellthird{5.67} & \cellthird{0.129} & \cellthird{0.135}  & \cellsecond{5.02} & \cellthird{20.90} & \cellthird{0.114} & \cellthird{7.84} & \cellthird{0.173} & 0.170 & \cellfirst{0.433} \\
- After Block-11        & \cellsecond{0.182} & \cellsecond{5.92} & \cellsecond{0.138} & \cellsecond{0.140}  & \cellthird{5.30} & \cellsecond{20.14} & \cellsecond{0.122} & \cellsecond{8.52} & \cellsecond{0.184} & \cellsecond{0.191} & \cellthird{0.382} \\
- After Block-22      & 0.146 & 4.29 & 0.120 & 0.123 & 6.90  & 25.20 & 0.098 & 6.83 & 0.168 & \cellthird{0.174} & 0.37 \\

- Interleaved (ours)    & \cellfirst{0.192} & \cellfirst{6.53} & \cellfirst{0.150} & \cellfirst{0.155}   & \cellfirst{4.79} & \cellfirst{18.91} & \cellfirst{0.131} & \cellfirst{9.21} & \cellfirst{0.210} & \cellfirst{0.222} & \cellsecond{0.415}   \\

\midrule
\emph{Feature injection:} &&&& \\
- Concat. to text w/FT     & \cellsecond{0.186} & 4.12 & \cellsecond{0.124} & \cellsecond{0.128} & \cellfirst{3.36} & \cellsecond{20.35} & 0.100 & \cellfirst{9.56} & \cellthird{0.186} & \cellthird{0.196} & 0.355 \\
- Direct alignment         & 0.098 & 2.38 & 0.029 & 0.030 & 8.16 & 42.35 & 0.065 & 3.90 & 0.094 & 0.100 & 0.283  \\
- Direct alignment w/FT    & 0.110 & 3.15 & 0.028 & 0.030 & 9.26 & 36.51 & 0.07 &  6.45 & 0.129 & 0.137 & 0.257  \\
- w/o symm. feature reinj. & 0.120 & \cellthird{4.57} & 0.059 & 0.062 &  \cellthird{6.60} & 30.27 & \cellthird{0.102} & 6.17 & 0.136 & 0.143 & \cellthird{0.365}   \\
- Symm. cross attention    & \cellthird{0.170} & \cellsecond{5.70} & \cellthird{0.118} & \cellthird{0.123}  & 8.47 &  \cellthird{22.30} & \cellsecond{0.126} & \cellthird{8.09} & \cellsecond{0.194} & \cellsecond{0.210} & \cellsecond{0.410}  \\
- Fusion blocks (ours)     & \cellfirst{0.192} & \cellfirst{6.53} & \cellfirst{0.150} & \cellfirst{0.155}  & \cellsecond{4.79} & \cellfirst{18.91} & \cellfirst{0.131} & \cellsecond{9.21} & \cellfirst{0.210} & \cellfirst{0.222} & \cellfirst{0.415}   \\

\bottomrule

\end{tabular}
}
\end{center}
\caption{\label{tab:config_comparison} V2A ablation results of our method. Variants marked with \emph{FT} indicate backbone finetuning when few parameters are introduced. }
\end{table*}

This section first introduces datasets (see \secref{sec:datasets}) and evaluation protocol (see \secref{sec:evaluation_protocol}). We then and compare our method to the state-of-the-art in V2A and A2V in \secref{sec:baselines_evaluation} and present an analysis of the proposed components in \secref{sec:ablations}. We present extensive qualitative results supporting our evaluation in \apref{sec:additonal_evaluation} and the \website{}.

\subsection{Datasets}
\label{sec:datasets}
We train the base video model on an internal automatically captioned video dataset and the base audio model on ambient and music sound clips sourced from AutoReCap and AutoReCap-XL \cite{genau}, FMA \cite{fma_dataset}, Magna \cite{magna_dataset}, MTG \cite{mtg_dataset}, and Song Describer \cite{sogndescriber}. The Fusion blocks are trained on 200,000 videos from VGGSounds~\cite{chen2020vggsound} and the temporally-strong AudioSet~\cite{gemmeke2017audioset} dataset. VGGSounds is an audio-visual dataset containing 126,000 samples. The temporally-aligned AudioSet includes 103,000 audio-video clips with strong audio-visual alignment. Additionally, we finetune the model on a high-quality internal captioned audio-video dataset comprising 250,000 clips. Audio captions are generated using the AutoCap \cite{genau} audio captioning model.

\begin{figure*}[t]
    \centering
    \includegraphics[width=\linewidth]{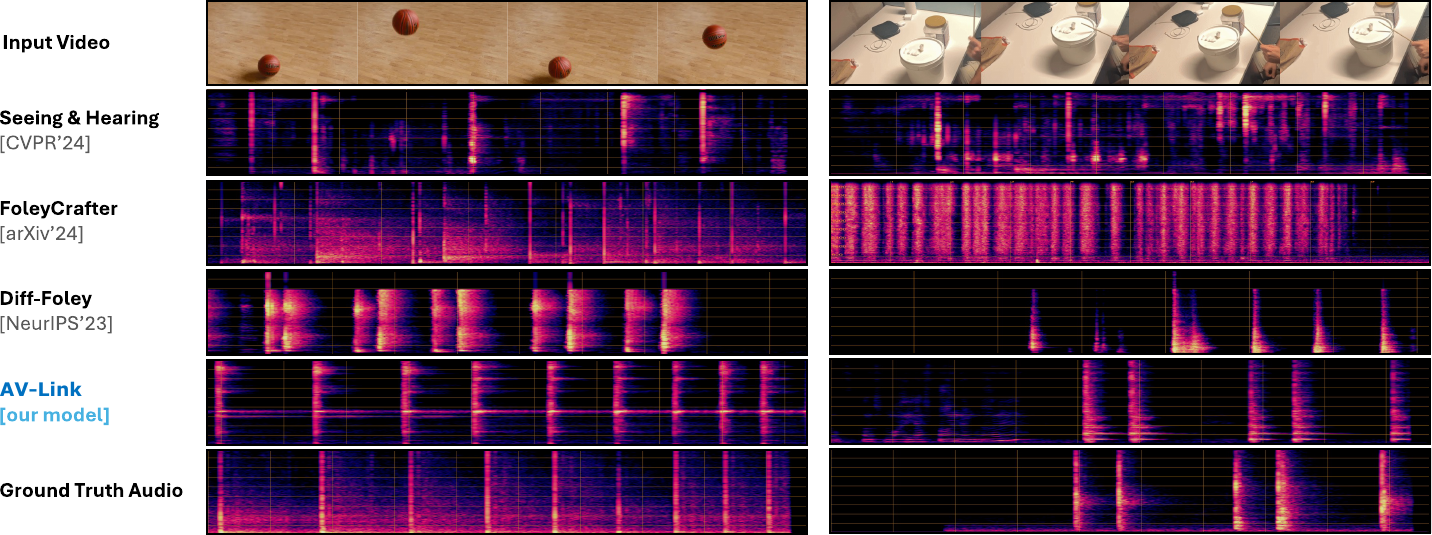}
    \caption{Qualitative V2A results. Our model achieved the best temporal alignment, matching closely the ``bouncing'' and ``drumming'' sounds entailed by the video modality. See the \supp{} and \website{} for additional results.}
    \label{fig:v2a_qualitatives}
\end{figure*}

\subsection{Evaluation Protocol}
\label{sec:evaluation_protocol}
We separately evaluate our model on the V2A and A2V tasks. We consider a conditioning video or audio input from the test set and sample a corresponding audio or video using our V2A or A2V model respectively. As prior work does not consistently adopt conditioning text prompts at inference~\cite{difffoley,foleycrafter,moviegen}, we report results for both settings. In all experiments, we use 8 fusion blocks (186M parameters), a learning rate of $1e\!-\!4$, and a batch size of 128 on 8 A100 GPUs for the ablation, increasing to 256 on 16 A100 GPUs for final experiments.

\noindent\textbf{Baselines.} For the task of V2A, we compare with Diff-foley~\cite{difffoley}, FoleyCrafter~\cite{foleycrafter}, Seeing and Hearing~\cite{seeingandhearing} (S\&H), Frieren~\cite{frieren}, V2A Mapper~\cite{wang2023v2amapper} and Movie Gen~\cite{moviegen}. Diff-foley and V2A-Mapper conditions an audio generator on CAVP and CLIP, respectively by concatenating it with text embeddings. FoleyCrafter adapts a pre-trained audio generator for A2V using a semantic adapter with frame-based CLIP features~\cite{clip} and trains a temporal adapter that aligns audio based on an onset mask predicted from the video, inspired by ControlNet. S\&H uses a training-free approach that optimizes cosine similarity in ImageBind~\cite{imagebind} between video and generated audio. Movie Gen and Frieren proposed to concatenate frame-wise features with audio tokens. For A2V, we compare with TempoToken \cite{tempotoken}, which optimizes latent tokens on a pre-trained video generator using BEATs~\cite{beats} audio features.

\begin{table}[t!]
\centering
\resizebox{\linewidth}{!}{%
\begin{tabular}{l c c c c c}
\toprule
 & Prompt & FID $\downarrow$ & $\text{FVD}_{12}$ $\downarrow$  & IB-AI $\uparrow$ & IB-AV $\uparrow$  \\
\midrule
TempoTokens~\cite{tempotoken} & & \cellthird{81.43} & \cellthird{1489.35} & \cellthird{0.112} & \cellthird{0.114}   \\
Ours-Joint & & \cellsecond{51.63} & \cellsecond{297.24} & \cellsecond{0.112} & \cellsecond{0.126} \\
Ours & & \cellfirst{39.10} & \cellfirst{266.93} & \cellfirst{0.160}  & \cellfirst{0.168} \\
\midrule

TempoTokens~\cite{tempotoken} & \checkmark & \cellthird{77.87} & \cellthird{1275.50} & \cellthird{0.167} & \cellthird{0.173} \\
Ours-Joint & \checkmark & \cellsecond{40.72} & \cellsecond{206.30} & \cellsecond{0.185} & \cellsecond{0.192} \\
Ours & \checkmark &  \cellfirst{37.62} &  \cellfirst{158.71} &  \cellfirst{0.191} & \cellfirst{0.201}  \\
\bottomrule
\end{tabular}}
\caption{\label{tab:baselines_comparison_a2v} Comparison to baselines on the A2V task on VGGSounds~\cite{chen2020vggsound}.}
\vspace{-2mm}
\end{table}

\noindent\textbf{Metrics} 
We follow the protocol of previous work~\cite{difffoley, tempotoken, seeingandhearing}  and adopt 2048 videos from the VGGSounds~\cite{chen2020vggsound} test set. 
We employ Fréchet Audio Distance (FAD)~\cite{audioldm}, Inception Score (IS)~\cite{salimans2016improved} and  CLAP~\cite{wu2023laionaudio} score as measures for audio quality and prompt following. We also report FID~\cite{heusel2017gans} and FVD~\cite{unterthiner2018towards} as measures of video generation quality for the A2V task. Additionally, we leverage ImageBind~\cite{imagebind} similarity scores between audio-image (IB-AI) and audio-video (IB-AV) modalities to measure semantic alignment. Onset detection accuracy (Onset ACC)~\cite{conditionalgenerationaudiovideo} is used as the metric for measuring temporal alignment for the V2A task. 
For user studies, we ask evaluators to express preferences for one of two results based on \emph{Audio Quality}, \emph{Video Quality}, paired \emph{Audio-Video Quality}, \emph{Semantic Alignment} and \emph{Temporal Alignment} between the two modalities. We report full details on user studies in \apref{sec:additonal_evaluation}. Furthermore, We adopt the newly introduced Movie Gen V2A sound effects Benchmark~\cite{moviegen}, consisting of 527 \emph{generated} video samples, for ablation studies and evaluating against Movie Gen. Note that since ground truth audio is not available for this benchmark, we omit audio-aligned metrics.

\subsection{Results}
\label{sec:baselines_evaluation}
\noindent\textbf{Video-to-Audio} We compare \method{} with baseline approaches on the V2A task in \tabref{tab:baselines_comparison_v2a}. Our method achieves the highest scores across all metrics, except for IB-AI and IB-AV, where S\&H reports superior values. This is expected, as S\&H explicitly optimizes for these metrics as part of its approach, making these scores less indicative of overall quality. Notably, our method demonstrates significantly better temporal alignment, with an Onset Accuracy of 0.53 compared to the top baseline score of 0.3. Such improvements are evident with and without providing audio text descriptions. For a more balanced comparison, we also evaluate our model trained only on VGGSounds, showing that it still substantially outperforms baselines.
This advantage is further confirmed by our user studies in \tabref{tab:user_study_baselines_unified}, where \method{} is consistently preferred, particularly for temporal alignment. Additionally, while Movie Gen achieves better sound quality due to its larger model (13B parameters), our method's audio-video temporal alignment is preferred 63.6\% of the time. 
We also illustrate our model's strong temporal alignment in \figref{fig:v2a_qualitatives} and on the \website{}, where we perform inference on videos requiring precise temporal audio correspondence such as `bouncing' and `tapping' sounds. While baselines capture semantic content, they fall short in synchronizing the precise onset of these events.

\begin{table}[t!]
\begin{center}
\setlength{\tabcolsep}{2.0pt}
\footnotesize
\resizebox{\linewidth}{!}{%
\begin{tabular}{l c | c c c c c}

\toprule

Configuration & Prompt & \emph{A-Qual.} & \emph{V-Qual.} & \emph{AV-Qual.} & \emph{Sem. Align.} & \emph{Temp. Align} \\

\midrule

\emph{Video-to-Audio:}  &  &  &  & \\
-Diff-Foley~\cite{difffoley} & & \cellgreendark{78.0} & - & \cellgreendarkdark{86.1} & \cellgreendark{84.9} & \cellgreendark{83.7}  \\
-V2A-Mapper~\cite{wang2023v2amapper} &  & \cellgreen{64.8} & - & \cellgreen{66.8} & \cellgreen{65.2} & \cellgreen{71.6} \\
-Frieren~\cite{frieren} &  & \cellgreen{66.4} & - & \cellgreen{64.4} & \cellgreen{63.2} & \cellgreen{70.4} \\
-Seeing and Hearing~\cite{seeingandhearing} & & \cellgreendarkdark{86.5} & - & \cellgreendarkdark{97.1} & \cellgreendarkdark{95.1} & \cellgreendarkdark{95.5} \\
-Seeing and Hearing~\cite{seeingandhearing} & \checkmark  & \cellgreendark{76.2} & - & \cellgreendarkdark{87.7} & \cellgreendarkdark{86.8} & \cellgreendarkdark{88.1} \\
-FoleyCrafter~\cite{foleycrafter} &  & \cellgreen{66.5} & - & \cellgreendark{76.3} & \cellgreendark{75.5} & \cellgreendark{80.0} \\
-FoleyCrafter~\cite{foleycrafter} & \checkmark & \cellgreenlight{57.4} & - & \cellgreen{64.3} & \cellgreen{67.2} & \cellgreen{65.5} \\
-Movie Gen~\cite{moviegen} & \checkmark & \cellred{34.4} & - & \cellgreenlight{52.8} & \cellgreenlight{56.8} & \cellgreen{63.6} \\

\midrule

\emph{Audio-to-Video:}  &  &  &  &  \\
-TempoTokens~\cite{tempotoken} &  & - & \cellgreendarkdark{75.5} & \cellgreendark{78.0} & \cellgreen{73.9} & \cellgreendark{74.5} \\
-TempoTokens~\cite{tempotoken} & \checkmark & - & \cellgreendarkdark{96.2} & \cellgreendark{83.8} & \cellgreendark{75.2} & \cellgreen{76.2} \\

\bottomrule

\end{tabular}
}
\end{center}
\caption{\label{tab:user_study_baselines_unified} User study comparing \method{} against baselines. Results in \% of votes in favor of our method.}
\vspace{-2mm}
\end{table}

\begin{figure}[t]
    \centering
    \includegraphics[width=\linewidth]{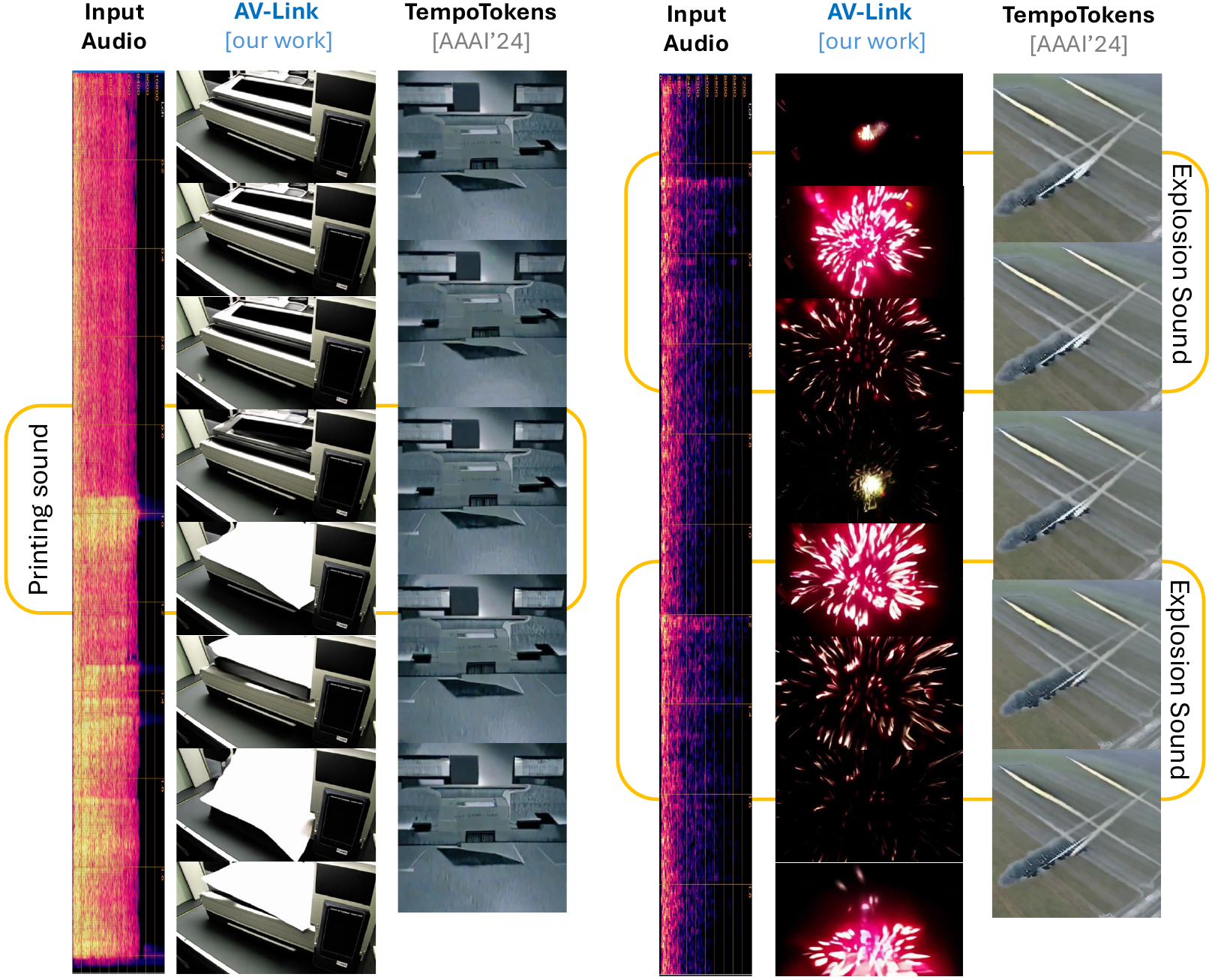}
    \caption{Qualitative results of A2V generation. Our model generates semantically and temporally aligned content capturing temporal events implied by the audio modality such as explosions and printing. See the \website{} for additional results.}
    \vspace{-2mm}
    \label{fig:a2v_qualitatives}
\end{figure}

\noindent\textbf{Audio-to-Video} On the A2V task, our method surpasses TempoTokens~\cite{temporallyalignedaudiovideo} in both video quality and audio-video alignment, as shown in~\tabref{tab:baselines_comparison_a2v}. User studies in~\tabref{tab:user_study_baselines_unified} also indicate that our approach is preferred for generation quality, semantic, and temporal alignment, with a preference greater than $73.9\%$ across all settings. Additionally, qualitative results in ~\figref{fig:a2v_qualitatives} and on the \website{} demonstrate that our method produces videos that align closely with the audio signal, both semantically and temporally. Our generated video events (e.g., explosions) occur precisely in sync with corresponding audio events, showcasing the strength of our \method{} in maintaining temporal coherence.

\subsection{Ablation studies}
\label{sec:ablations}
This section evaluates the contribution of each method's design choices to output quality. Unless otherwise specified, we employ frozen audio and video generators, and focus on the V2A task, whose findings we expect to generalize to A2V due to the symmetric nature of our approach. We report the main ablation results in \tabref{tab:config_comparison} for models trained on VGGSounds dataset. To obtain more accurate qualitative and quantitative comparisons between baselines, we perform evaluation using a single fixed seed for all test videos.

\noindent\textbf{Which is the optimal conditioning flow timestep?}
Diffusion models learn distinct features at different layers and for different diffusion timesteps~\cite{ddpmseg, openvocabularypanopticsegmentationtexttoimage, hyperfeatures}. Thus, the choice of flow time distribution for the conditioning modality $\difftimestep\sim\timestepdistribution$ determines the quality of the conditioning features. Taking the V2A example, using a high-noise flow timestep $\difftimestepvideo$ would destroy the video signal, leading to loss of conditioning information, while a flow time $\difftimestepvideo\!=\!1$ (\ie with no noise) would produce uninformative video activations, as for $\difftimestepvideo\!=\!1$, Eq.~\eqref{eq:rf} is minimized when the generator function is the identity $\generatorvideo(\inputvideoclean,1)\!=\!\inputvideoclean - \mathbb{E}_{\inputvideonoise\sim\noisedistribution\!=\!\mathcal{N}(0,\mathbb{I})}[\inputvideonoise]\!=\!\inputvideoclean$. 
We study the optimal choice of $\difftimestepvideo$ empirically by training a generator $\generatorvideotoaudio$ where the video time is uniformly sampled $\difftimestepvideo\!\sim\!\mathcal{U}(0,1)$ during training. At inference, we conduct evaluation for different values of $\difftimestepvideo$ and notice that the optimal value is close to a fully denoised video (see \figref{fig:video_time_choice}). Intuitively, small changes in the video can determine large variations in the sound, such as smoke from a gunshot, thus the model benefits from a noise level where high-frequency details are still not erased by noise. A symmetric analysis in \figref{fig:video_time_choice} shows a similar behavior for $\difftimestepaudio$ in the A2V task.
In addition, training with a uniform distribution of $\difftimestep$ for the conditioning modality rather than fixing it at its optimal value led to a slower convergence and reduced performance (see \tabref{tab:config_comparison}). Therefore, after identifying the optimal conditioning timestep to be 0.96 for V2A and 0.8 for A2V, we train subsequent models using these fixed timesteps.

\noindent\textbf{Can V2A and A2V models share the same parameters?}
We jointly train V2A and A2V models by sharing Fusion Block parameters of the two tasks. As reported in \tabref{tab:baselines_comparison_v2a} and \tabref{tab:baselines_comparison_a2v}, using separate parameters contributes to a marginal but consistent improvement over parameter sharing.

\noindent\textbf{Can both modalities be generated at the same time?}
\figref{fig:video_time_choice} suggests a key issue with joint video-audio diffusion:  Audio often depends on element that becomes only visible at the end of the sampling trajectory. For instance, accurately generating the sound of a gunshot requires high-frequency visual details, such as the smoke of an explosion, Similarly, sounds like thunder and explosions only become distant at the later sampler steps, making jointly denoising the two modalities challenging. This may explain the lower performance of current methods in this task~\cite{mmdiffusion,seeingandhearing,avdit}, and suggest that audio-video generation benefits from factorization into video generation followed by V2A and vice-versa. Indeed, when we train \method{} for joint audio-video generation, we observed a substantial drop in performance.

\noindent\textbf{Are Fusion Blocks necessary?}
We evaluate several methodologies for injecting conditioning features into the model for which we show results in the \emph{Feature injection} section of \tabref{tab:config_comparison}. \emph{Concat. to text} follows Diff-foley \cite{difffoley} in concatenating conditioning features to the text embeddings. \emph{Direct alignment} follows Movie Gen \cite{moviegen} in repeating the visual features to match the number of audio frames and sum them together. \emph{Symm. cross attention} replaces the symmetric design of Fusion Blocks based on self attention with two separate blocks based on cross attention: the first for injecting features into the audio backbone, the second for the video. The Fusion Block design consistently outperforms the baselines. Additionally, we explore using video backbone as a static feature extractor without reinjecting the output back to the subsequent video DiT blocks. While this approach allows for caching the video features and thus improves sampling efficiency, it significantly underperforms our method that progressively refines the conditioning features by reinjecting them back to subsequent DiT blocks.

\noindent\textbf{How do diffusion features compare to pretrained representations?}
We experimented with various pretrained feature extractors in place of frozen generator activations as inputs to the Fusion Blocks. As shown in \tabref{tab:config_comparison}, diffusion features outperform all representations on the Movie Gen benchmark. On VGGSounds it shows the strongest temporal alignment as expressed by Onset ACC. User studies against baselines employing such representations (see \tabref{tab:user_study_baselines_unified}) confirm improved temporal and semantic alignment.

\noindent\textbf{Which layers yield the best conditioning features?}
In \tabref{tab:config_comparison}, we ablate over four different designs where the multimodal Fusion Blocks are placed at early video DiT layers, in the middle, in the end, or regularly interleaved with the video DiT blocks. Interleaving the Fusion Blocks outperforms the other variants, indicating that feature diversity and regular modality alignment benefit the model.

\section{Conclusion}
\label{sec:conclusion}

We present \method, a unified approach for V2A and A2V generation. It leverages time-aligned activations from pretrained flow models, bypassing the use of specialized feature extractors. To enable this, we introduce a Fusion Block that allows for bidirectional information exchange between frozen generators for each modality through the use of temporally-aligned self attention and symmetric feature reinjection. Extensive ablations and comparisons show performance improvements over all metrics, with particular regard to temporal alignment. \\
\textbf{Acknowledgements.} The authors thank Maryna Diakonova and Ke Ma for their help in preparing the data and user studies. We also thank Chaoyang Wang, Sherwin Bahmani, Tsai-Shien Chen and Ziyi Wu for their guidance.

{
    \small
    \bibliographystyle{ieeenat_fullname}
    \bibliography{main}
}

\clearpage
\setcounter{page}{1}
\appendix
\maketitlesupplementary

\noindent We discuss limitations in Sec.~\ref{sec:ap_limitations}. We then include details on the architecture (Sec.~\ref{sec:architecture_details}), training and inference (Sec.~\ref{sec:training_details}), and evaluation (Sec.~\ref{sec:additonal_evaluation}). We \emph{highly} encourage the reader to visit the attached \website{} for extensive qualitative results and comparisons.

\section{Limitations}
\label{sec:ap_limitations}

Our base video backbone is a low-resolution and low-fps RGB model coupled with a high-resolution upsampler model. While the model achieves state-of-the-art V2A and A2V performance, leveraging a large high-resolution latent video model may further improve performance. Exploring model scaling to further improve feature quality is an exciting avenue for future work. Additionally, the feature reinjection alters activations in the conditioning modality generator. While beneficial, it introduces additional compute at each sampling step as caching prior activations becomes infeasible. Step distillation techniques reduce this effect by reducing the number of sampling steps and constitute an orthogonal line of work.

\section{Architecture Details}
\label{sec:architecture_details}
We base our audio and video backbones on a shared DiT~\cite{dit} architecture.
For the video model, We use an pixel-based flow matching model with an initial patchification operation of $2\times2$ using a patch dimension of 1024. We employ 24 DiT blocks. Each DiT block is composed of a self attention operation, followed by a cross attention operation attending to text conditioning signals, and a final MLP. We use 16 heads for each attention operation and a hidden dimension of 4096 for the final MLP. Adaptive layer normalization is used within each block to condition the model on the flow time $t$. Each attention operator makes use of QK-Normalization~\cite{sd3} to improve training stability when trained in BF16 precision, and 3D RoPE~\cite{rope} positional embeddings.

For the audio model, we employ a latent model with 24 DiT blocks with hidden dimension of 1024 for the patches. We use 16 attention heads for each self-attention and cross-attention operation and an MLP hidden size of 4096. We follow \cite{genau} for encoding the audio and converting the generated Mel-spectrograms to waveform. Both models have 576M trainable parameters.

The Fusion blocks similarly have a hidden dimension of 1024 and 16 heads for the self-attention operation. Their final MLP layers have a hidden dimension of 4096. Each Fusion block has 23.25M parameters and we use 8 Fusion blocks for all of our experiments. 

\section{Training and Inference Details}
\label{sec:training_details}

This section presents additional details on training and inference and discusses training and inference time.

\subsection{Training Details}
For all training phases, we train our models using the AdamW optimizer with a learning rate of 3e-4, beta factors of 0.9 and 0.99, epsilon of 1e-8, a weight decay of 0.01, and a 10,000-step warmup. 

The base video model is trained for 250,000 steps on an automatically-captioned internal dataset with a total batch size of 512 on 16 A100 GPUs. The base audio models are trained for 100,000 with a total batch size of 1024 on 8 A100 GPUs. We drop text condition 10\% of the time to enable classifier-free guidance (CFG)~\cite{cfg} during inference.

The fusion blocks are trained for 50,000 steps on 16 A100 GPUs with a batch size of 256. Ablation experiments are trained on 8 A100s with a total batch size of 128. We drop the generated modality text prompt (\eg audio text prompt in V2A task) 50\% of the time and the conditioning modality text prompt (\eg video text prompt in V2A task) 20\% of the time. For all of our experiments, both the audio and video backbone are kept frozen unless otherwise specified. 

\subsection{Inference details}
We perform inference starting from pure Gaussian noise for the modality to generate and use the model's velocity estimates together with an Euler sampler to progressively transform the noise to the clean generated sample. We found using classifier-free guidance on the conditioning modality to be instrumental in obtaining good multimodal alignment. When conditioning on more than one modality (\eg video and audio text prompt), we drop both conditions simultaneously to compute the unconditional signal. In all of our experiments, we used 64 sampling steps and a CFG weight of 5.0.  All of our generated video results are evaluated at the 512 x 288 pixels resolution.

\subsection{Training and Inference Time}
\textbf{Training.} The base video model was trained for 25 days, while the audio model and the fusion blocks were trained for 8 and 4 days, respectively. All experiments utilized PyTorch Fully Sharded Data Parallel (FSDP)~\cite{fsdp} for efficient distributed training.

\noindent\textbf{Inference.} We measure a throughput of 27.85s per sample using a batch size of one on an A100 to perform 64 sampling steps for both the A2V and V2A tasks. This is a limitation of our method in the V2A task since it results in a slower sampling time compared with previous approaches~\cite{difffoley, foleycrafter}. We make, however, two important considerations. First, a major use case for V2A is the sonification of generated soundless videos. Such generation, when performed by state-of-the-art large-scale text-to-video generators usually takes several minutes. Second, since our method relies on a flow model, distillation methods similar to the ones adopted by previous approaches~\cite{frieren} can significantly improve inference time by performing sampling in a few steps. We regard this direction as an interesting avenue for future work that is orthogonal to ours.

\section{Additional Evaluation Results and Details}
\label{sec:additonal_evaluation}

This section presents additional evaluation results and details. We highly encourage the reader to experience the generated audios and videos on the \website{}.

\subsection{Noise Sampling Scheduler}

\begin{figure}
    \centering
    \includegraphics[width=0.99\linewidth,clip]{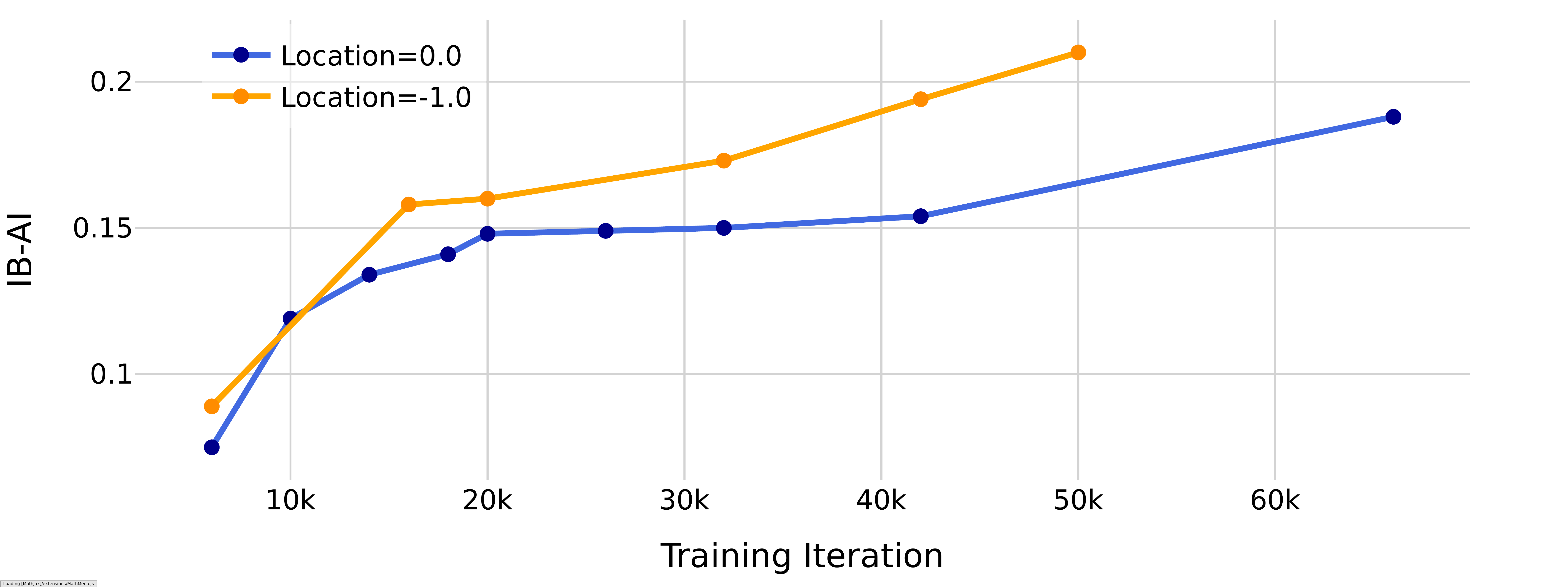}
    \caption{Comparison between different parametrizations for the Logit-Normal training distributions $\timestepdistribution$ for the flow timestep $\difftimestep$. When the location (\ie the mean of the normal distribution) is shifted towards higher noise levels, we observe faster model convergence.}
    \label{fig:convergance_speed}
\end{figure}

For training the video and audio model, we use the Logit-Normal distribution $\timestepdistribution$ for the flow timestep $\difftimestep$ with location 0.0 and scale 1.0, following~\cite{sd3}.
When training the Fusion block, which builds on pretrained audio and video backbones, we adopt a different approach. We noticed that early diffusion steps are the most critical for correctly following the conditioning modalities. Toward the end of the sampling path, however, the model relies more on the generated modality signal alone. Therefore, since multimodal alignment is the main purpose of the Fusion blocks, we shift the flow time training distribution to sample from more noisy steps. More specifically, Instead of parametrizing the normal distribution component as $\mathcal{N}(0, 1)$, we set it to $\mathcal{N}(-1, 1)$ and observe significantly faster convergence as shown in \figref{fig:convergance_speed}.

\subsection{Baselines Selection}
Below, we include details on baseline selection and inference procedures.

\noindent\textbf{Video-to-Audio.} We compare our method against Diff-Foley~\cite{difffoley}, FoleyCrafter~\cite{foleycrafter}, Seeing and Hearing~\cite{seeingandhearing}, Frieren~\cite{frieren}, and Movie Gen A2V~\cite{moviegen}.
\begin{itemize}
    \item \textbf{Frieren:} We use the Frieren (reflow) model, 64 sampling steps, CFG of 5.0, and the other recommended hyper-parameters for inference. 
    \item \textbf{Movie Gen A2V:} Since Movie Gen is a closed source model and audio samples are released for the Movie Gen Benchmark only, we include user studies and extensive comparison on the \website{} comparing our method against the released benchmark, showing that \method{} achieves superior temporal alignment.
    \item \textbf{FoleyCrafter:} We employ FoleyCrafter with default settings for inference.
    \item \textbf{Seeing and Hearing:} We use the official code for V2A and we follow the default sampling parameters. We exclude Seeing and Hearing qualitative comparison without prompt as it produces barely audible sounds in this setting. 
    \item \textbf{Diff-Foley:} We use the official code and set sampling steps to 64 and CFG to 5.0. For the rest of the parameters, we use the default setting. 
\end{itemize}
For all of the baselines, we generate the audio at their recommended length from the full-length videos and crop it to 5.16s for a fair comparison with our method. 

\noindent\textbf{Audio-to-Video.} To the best of the authors' knowledge, TempoToken~\cite{tempotoken} is the only in-the-wild A2V baseline with publicly available code. We exclude Seeing and Hearing~\cite{seeingandhearing} A2V results as their code is not available for this task. Additionally, we exclude joint audio-video generation methods (see \apref{sec:jointaudiovideo}) such as MMDiffusion~\cite{mmdiffusion}, as they were trained on the very limited Landscapes Videos dataset~\cite{lee2022soundguidedsemanticvideogeneration}, which contains only 928 videos and lacks generalization beyond landscape scenarios. We also exclude sound-guided image animation methods~\cite{audiosynchronizedvisualanimation, lee2022soundguidedsemanticvideogeneration}, as they address a fundamentally different task.

We use the official implementation of TempoToken with default parameters to generate 2-second videos. For the quantitative comparison, we crop our generated videos to 2 seconds for a fair comparison with TempoToken.

\subsection{Joint Audio-Video Generation Baselines}
\label{sec:jointaudiovideo}

In this work, we aim to address the tasks of A2V and V2A generation within a single framework. Some methods for joint video-audio generation~\cite{discriminatorguidedcooperativediffusionjoint,versatilediffusiontransformermixture, cmmdcontrastivemultimodaldiffusion, languageguidedjointaudiovisualediting, sequentialcontrastiveaudiovisuallearning,visionaudiobeyondunified} are capable of operating under this conditional setting.

However, due to the difficulties in jointly modeling the audio and video modalities, these methods are often trained on domain specific datasets: Landscapes~\cite{lee2022soundguidedsemanticvideogeneration} is composed of 928 videos of natural landscapes; AIST++~\cite{li2021learn} contains 1020 clips (5.2 hours) of dancing human sequences; GreatestHits~\cite{owens2016visuallyindicated} is composed of 977 videos featuring a drumstick hitting objects in a scene; EPIC-SOUNDS~\cite{huh2023epicsounds} encompasses 117.6k clips (100 hours) of cooking-related actions; Monologues~\cite{versatilediffusiontransformermixture} features 19.1M clips of talking people.
The use of narrow distribution datasets coupled with limited availability of source code and pretrained checkpoints limits the possibility of performing meaningful comparisons on the broad data distribution modeled by~\method{}.

\begin{figure*}
    \centering
    \includegraphics[width=\linewidth]{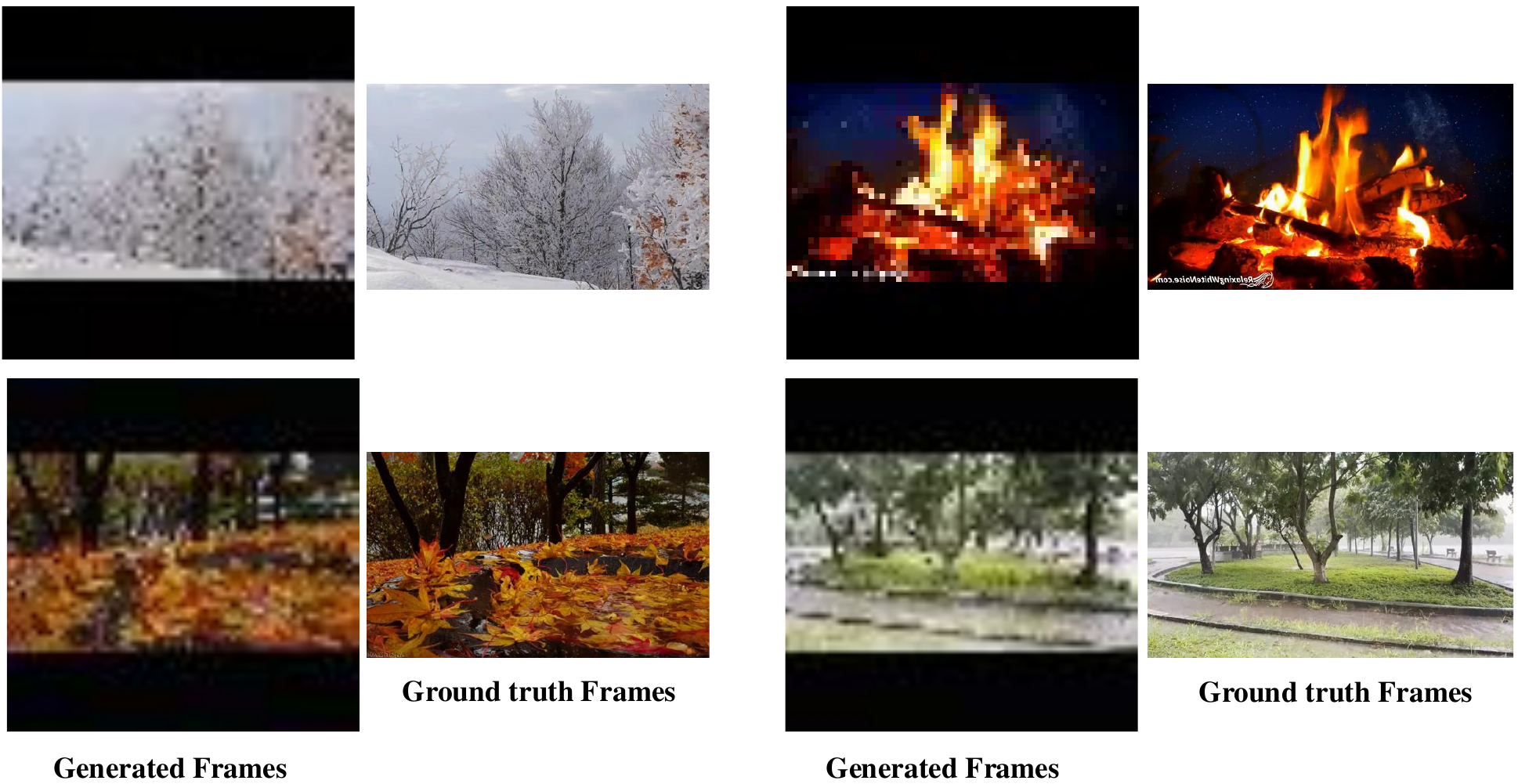}
    \caption{Examples MMdiffusion generated samples using their released checkpoint. We show that their model suffers from severe overfitting due to training on a limited dataset of 900 landscape videos.}
    \label{fig:mmdiffusion_overfitting}
\end{figure*}

In the following, we discuss the considered joint audio-video baseline methods. 
MMDiffusion~\cite{mmdiffusion} provides checkpoints for models trained on Landscapes~\cite{lee2022soundguidedsemanticvideogeneration} and AIST++~\cite{li2021learn} datasets only. We find that the Landscapes checkpoint overfits to its training dataset distribution, frequently replicating training samples. By analyzing 500 generated videos using the provided checkpoint, we found that the generated videos achieved an average of 0.825 CLIP similarity between the generated videos and the top-matching video from the training dataset. We include in \figref{fig:mmdiffusion_overfitting} examples of such overfitting.
This prevents the method from operating on videos outside these domains and does not allow for an informative comparison. 
Ishii \etal~\cite{simplestrongbaselinesounding} provides a checkpoint trained on the GreatestHits~\cite{owens2016visuallyindicated} dataset only. While trained also on the Landscapes~\cite{lee2022soundguidedsemanticvideogeneration} and VGGSound~\cite{chen2020vggsound} datasets, no A2V or V2A results were reported.
Kim \etal~\cite{versatilediffusiontransformermixture} report results on the Landscapes~\cite{lee2022soundguidedsemanticvideogeneration}, AIST++~\cite{li2021learn} and the Monologues (talking heads) datasets with no code publicly available.
CMMD~\cite{cmmdcontrastivemultimodaldiffusion} report results on the AIST++~\cite{li2021learn} and EPIC-SOUNDS~\cite{huh2023epicsounds} datasets and do not provide code.

\subsection{Additional V2A results}
\begin{figure*}[t]
    \centering
    \includegraphics[width=\linewidth]{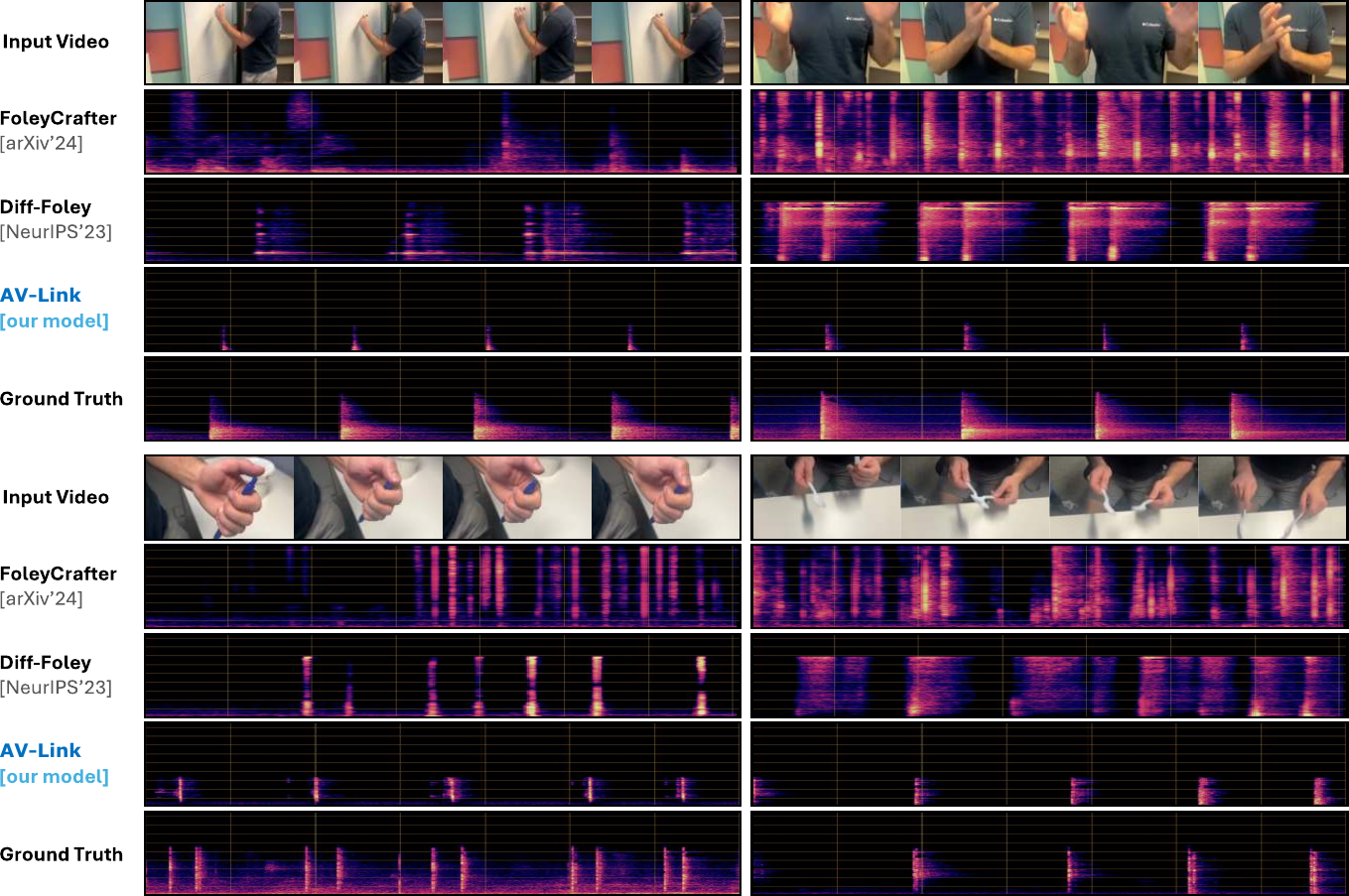}
    \caption{Qualitative V2A results comparing our method to baselines on in-the-wild videos captured by the authors that require precise temporal alignment. \method{} produces audio signals that closely align to the visual modalities, while baselines often produce audio that is unrelated or not correctly synchronized with the visual content. See the \website{} for more results.}
    \label{fig:v2a_qualitatives_supplement}
\end{figure*}

\figref{fig:v2a_qualitatives_supplement} shows additional V2A results comparing our method to baselines. To better showcase the capabilities of \method{}, we record a series of in-the-wild videos that require a high degree of temporal alignment for the audio modality and run inference for all baselines. Our method produces highly aligned audio results that capture the audio semantic entailed by the visual modality, while baselines produce degraded results. We attribute this phenomenon to the lack of access to visual features that are precisely aligned with the video content. On the contrary, the use of activations from the video generation backbone allows \method{} to produce precise alignment in this scenario.

\subsection{User study details}

We hire a team of professional annotators to perform the user studies. A total of 50 samples are generated for each method in the V2A and A2V tasks. We present users with paired videos with accompanying audio generated by different methods, and ask them to express a preference for one of the two based on \emph{Audio Quality}, \emph{Video Quality}, paired \emph{Audio-Video Quality}, \emph{Semantic Alignment} and \emph{Temporal Alignment} between the two modalities. For the case of V2A evaluation, we formulate instructions for annotators for each such aspect as follows:
\begin{itemize}
\item \textbf{Audio Quality} \emph{``Which audio has the best quality? Only listen to the audio and ignore the video content.''}
\item \textbf{Audio-Video Quality} \emph{``Which audio-video pair has the best quality?''}
\item \textbf{Semantic Alignment} \emph{``Which audio is semantically closer to the video content?''}
\item \textbf{Temporal Alignment} \emph{``Which audio is more temporally aligned to the video content?''}
\end{itemize}
The formulation of questions for the A2V case is completely symmetric.
For each generated pair of samples, we ask 5 different users to express a preference to increase the robustness of the evaluation.

\end{document}